**Title:**

# Natural data structure extracted from neighborhood-similarity graphs

**One sentence summary:** Direct construction of sparse 'neighborhood-similarity' graphs overcomes many limitations of traditional high-dimensional data analysis.

**Authors:** Tom Lorimer[1*], Karlis Kanders[1] and Ruedi Stoop[1,2]

[1]Institute of Neuroinformatics, University of Zurich and ETH Zurich, Winterthurerstr. 190, 8057 Zurich, Switzerland.

[2]Institute for Computational Science, University of Zurich, Winterthurerstr. 190, 8057 Zurich, Switzerland.

*Correspondence to: lorimert@ini.ethz.ch

**Abstract**:

'Big' high-dimensional data are commonly analyzed in low-dimensions, after performing a dimensionality reduction step that inherently distorts the data structure. For the same purpose, clustering methods are also often used. These methods also introduce a bias, either by starting from the assumption of a particular geometric form of the clusters, or by using iterative schemes to enhance cluster contours, with uncontrollable consequences. The goal of data analysis should, however, be to encode and detect structural data features at all scales and densities simultaneously, without assuming a parametric form of data point distances, or modifying them. We propose a novel approach that directly encodes data point neighborhood similarities as a sparse graph. Our non-iterative framework permits a transparent interpretation of data, without altering the original data dimension and metric. Several natural and synthetic data applications demonstrate the efficacy of our novel approach.

**Main Text:**

Present day data typically consist of simultaneous measurements of many distinct characteristics. For the interpretation of data, high dimensional 'data spaces' pose a challenge, in particular because the data cannot be directly visualized (e.g., cytometry or neuroscience data *(1, 2)*). As a solution, data dimensionality reduction approaches are generally used (e.g., *(3, 4, 5)*). Such a step, however, inherently distorts the original distance relation between data points, which has an uncontrollable effect on subsequent low-dimensional analysis methods. As an alternative, clustering methods have been applied directly to high dimensional data, but these methods often use either an a priori model of cluster form (to define sets of 'similar' points), or use a model-free iterative optimization scheme based on local distances. Model-based algorithms bias the results by searching for, e.g., (mixtures of) Gaussian clouds or other convex sets *(6, 7, 8)*, manifold structures *(9)*, or regions of constrained density change *(10, 11)* that may not be inherent to the given data. Model-free approaches can avoid this shortcoming (and have been shown to produce strong results, e.g. *(12, 13)*), but the effects of iterative modifications of data point distances are generally not seizable, which may bias results in an unpredictable manner.

Our novel and general framework circumvents these problems. We build a sparse graph over the data, where data points comprise the nodes. Nodes are linked by edges only if they are close to each other and occupy structurally 'similar' regions of the dataset. In this way, the problem of interpreting data structure in high dimensions, is reduced to the problem of interpreting the structure of a sparse graph, for which a vast literature of methods is already available. Importantly, the time-complexity of these methods, when applied to our sparse graphs, will be largely independent of the original dimensionality of the dataset.

More precisely, we start from a directed *k*-nearest-neighbor (*k*NN) graph, defined with respect to a distance measure chosen to suit the particular data science application (often, Euclidean distance). The structural contours of this *k*NN graph are then sharpened by calculating a measure of 'structural similarity' of the pairs of data points joined by each edge, and removing (filtering) edges according to a user-defined thresholding criterion. To retain full control over the bias introduced by this edge filtering, we use the *full* distance information in the *k*-neighborhood of each point, independently, as follows. The set of distances from a given point to each of its *k*-nearest neighbors defines an empirical distribution of distance. This distribution can be quantitatively compared to the corresponding distributions of the neighboring points in the graph. The statistical literature contains many methods for comparing empirical distributions, each with well-documented and specific properties. As a proof of concept, we chose here the well-known Kolmogorov-Smirnov (K-S) statistic: The maximum absolute difference between two cumulative distributions (for our specific implementation, see Supplement). Two data points with empirical distance distributions that have a low K-S statistic, are likely to reside in structurally similar regions of the dataset. We emphasize, however, that any other quantitative measure of distribution similarity suited to meet the specific goals of the data scientist could be chosen. Nevertheless, simply comparing distance distributions in this way may be misleading (e.g., when two sections of a data manifold pass close to one another, or in other cases of 'symmetry' within the dataset). To remedy this shortcoming, we combine the latter measure with a measure of the overlap of the *k*-neighborhoods of the nodes joined by each edge in the *k*NN graph. Again, this can be done in many ways; we demonstrate the concept here using the Jaccard similarity index (see Supplement).

We first show how after filtering edges from the *k*NN graph as described above, natural data clusters can emerge as connected graph components, independently of any subsequent graph-based interpretation steps. To illustrate the strength of this approach, we begin with challenging two-dimensional examples, before adapting our approach further for high dimensional data. Spirals are a commonly used test challenge to explore the strength of new algorithms. Even beginning from a *k*NN graph with $k = 20$, it is possible to reveal by edge filtering the one dimensional chain structures of closely interwoven and sparsely sampled spirals (Fig. 1 B), without making an explicit manifold assumption. Moreover, because the similarity measures we use are based on comparative rather than absolute measures, they are largely density invariant, which permits edge removal *between* arbitrarily shaped high and low density regions, while preserving edges *within* them (Fig. 1 B).

For high-dimensional data, we propose a sorting of the nodes that exhibits the data structure at a wide range of scales visually, in the graph's adjacency matrix. Upon removing edges from the original *k*NN graph by means of edge filtering, we obtain nested sets of separate connected components. The order in which these natural topological clusters emerge, expresses the fundamental relationship between the data structure and the similarity criteria used for filtering. Algorithmically, the edges are filtered over a large sweep of an averaged value of the two similarity measures (see Supplement). Starting the sweep with the most strict edge filtering criterion (most edges removed), we sort the nodes according to the size of the (strongly) connected component to which they belong (Fig. 1 C). As the edge filtering criterion becomes more lenient, the connected components merge together, and the nodes are re-sorted according to the size of these new connected components, preserving the previous orderings as a sub-ordering (see Supplement). The process continues until the original *k*NN graph configuration is regained. While this procedure may look similar to hierarchical clustering, it is, in fact very different, as detailed neighborhood information is at the base of our approach, which permits to also successfully separate convex-concave sets *(14)*.

To validate the approach for high dimensions, we use a polynomial transformation that puts original low-dimensional data (over which we have perfect visual control) on a clear manifold structure in high-dimensional space (Fig. 2 A) *(15)*. While with such data, the celebrated t-SNE algorithm *(5)* at standard parameter settings has substantial difficulties (Fig. 2 B-D), our method is able to identify even the smallest, closely spaced sets (Fig. 2 E, F). This highlights a strength of the adjacency matrix sorting: In the absence of background noise, many of the major partitions are already captured in the *k*NN graph before filtering, so that this structure (and also its substructures) clearly emerge upon sorting (see Supplement for further details).

To scrutinize the performance on real-world high-dimensional data, two carefully curated standard image datasets, COIL-20 *(16)* and MNIST *(17)*, are examined. Both datasets contain labeled images of single objects (pre-processed to minimize the variation of presentation), where, rather than working with higher level features, which is what often is done with concurrent approaches, we take the pixel values themselves as the data. The COIL-20 dataset *(16)* contains images of 20 different objects photographed from different angles (128x128 grayscale pixels). Already in the sorted adjacency matrix, our approach successfully distinguishes the majority of the objects (Fig. 3). However, even after filtering, five of the objects remain within the same connected component. Closer inspection, however, reveals that, upon rotation, these five horizontal rectangular objects occupy the frame in a similar way (Fig. 3 D). Additionally, asymmetric object number 2 is partitioned into two distinct components, one for left facing and one for right, with front facing images associated with similarly shaped object number 8 instead (Fig. 3 D). This has a simple explanation: In our naive representation of the data space where each pixel value is a coordinate, the distances between light gray objects on a black background are dominated by the object's shape. The MNIST 'training' dataset *(17)* of 60000 images of handwritten digits (28x28 grayscale pixels) with its significant overlap between the digit classes *(15)* presents an even greater challenge. For these data, the connected component criterion for partitioning the graph is too strict to permit a breakdown into correct digit classes, so we demonstrate the potential of subsequent graph-based interpretation steps. By partitioning the largest connected component of the filtered graph using a standard spectral clustering approach (the Normalized Cut *(18)*), the digit classes can immediately be revealed from the graph topology (Fig. 4, see Supplement), where errors of the spectral clustering process (Fig. 4 B) can be remedied by removing the spurious partitions (Fig. 4 C). Here, the role of edge filtering is evident: Filtering removes the majority of the *k*NN graph edges between digit classes (Fig. 4 C), permitting the 10 dominant clusters to be revealed. Reassignment of the other nodes to their nearest large cluster (see Supplement) provides an accurate partition of the digit classes (Fig. 4 F, with F-measure 0.95). This compares favorably with other state-of-art unsupervised methods, such as InfoGAN *(19)* that, however, require that the number of digit classes is known, or request the training of a large neural network.

Our approach also offers a significant computational advantage: after the construction of the *k*NN graph, the edge-filtering operations can be performed at very low computational cost, so that even exceedingly large datasets can be analyzed by real-time interpretive exploration. The construction of the *k*NN graph, as the most computationally expensive step, can be accelerated by using approximate solvers (e.g., *(20)*). The only parameters that need to be set in our approach are the edge filtering thresholds (which can also be combined into a single parameter, as is done in the adjacency matrix sorting). We suggest that a value of *k* close to 20 (as was used for *all* examples presented here) will be sufficient for most datasets where the intrinsic dimension is not too large.

There is substantial scope for development of specific algorithms and transparent tools based on our general framework. The results presented here are only a small hint of the potential of this approach, suggesting a substantial role for neighborhood similarity graphs in the future of data science.

**References and Notes:**

1. S. C. Bendall, E. F. Simonds, P. Qiu, E. D. Amir, P. O. Krutzik, R. Finck, R. V. Bruggner, R. Melamed, A. Trejo, O. I. Ornatsky, R. S. Balderas, S. K. Plevritis, K. Sachs, D. Pe'er, S. D. Tanner, G. P. Nolan, Single-cell mass cytometry of differential immune and drug responses across a human hematopoietic continuum. *Science* **332**, 687-696 (2011).

2. J. N. Stirman, I. T. Smith, M. W. Kudenov, S. L. Smith, Wide field-of-view, multi-region, two-photon imaging of neuronal activity in the mammalian brain. *Nat. Biotechnol.* **34,** 857-862 (2016).

3. J. B. Tenenbaum, V. de Silva, J. C. Langford, A global geometric framework for nonlinear dimensionality reduction. *Science* **290**, 2319-2323 (2000).

4. S. T. Roweis, L. K. Saul, Nonlinear dimensionality reduction by locally linear embedding. *Science* **290**, 2323-2326 (2000).

5. L. van der Maaten, G. E. Hinton, Visualizing data using t-SNE. J. Mach. Learn. Research **9**, 2579-2605 (2008).

6. J. MacQueen, "Some methods for classification and analysis of multivariate observations" in *Proceedings of the Fifth Berkeley Symposium on Mathematical Statistics and Probability*, L. M. Le Cam, J. Neyman, Eds. (Univ. California Press, Berkeley, CA, 1967), vol. 1, pp. 281-297.

7. J. H. Ward Jr., Hierarchical grouping to optimize an objective function. *J. Am. Stat. Assoc.* **58**, 236-244 (1963).

8. C. M. Bishop, *Pattern recognition and machine learning* (Springer, New York, 2006) pp. 423-459.

9. R. Souvenir, R. Pless, "Manifold clustering" in Proceedings of the Tenth IEEE International Conference on Computer Vision (ICCV 2005), (IEEE, 2005).

10. M. Ester, H.-P. Kriegel, J. Sander, X. Xu, "A density-based algorithm for discovering clusters in large spatial databases with noise" in *Proceedings of the 2nd International Conference on Knowledge Discovery and Data Mining*, E. Simoudis, J. Han, U. Fayyad, Eds. (AAAI Press, Menlo Park, CA, 1996), pp. 226–231.

11. A. Rodriguez, A. Laio, Clustering by fast search and find of density peaks. *Science* **334**, 1492-1496 (2014).

12. M. Blatt, S. Wiseman, E. Domany, Superparamagnetic clustering of data. *Phys. Rev. Lett.* **76**, 3251-3254 (1996).

13. F. Landis, T. Ott, R. Stoop, Hebbian self-organizing integrate-and-fire networks for data clustering. *Neural Comput.* **22**, 273-288 (2010).

14. F. Gomez, R.L. Stoop, R. Stoop, Universal dynamical properties preclude standard clustering in a large class of biochemical data, *Bioinformatics* **30**, 2486-2493 (2014).

15. T. Lorimer, J. Held, R. Stoop, Clustering: how much bias do we need? *Phil. Trans. R. Soc. A* **375**, 20160293 (2017).

16. S. A. Nene, S. K. Nayar, H. Murase, "Columbia Object Image Library (COIL-20)" (Tech. Rep. CUCS-005-96, Columbia Univ., 1996).

17. Y. LeCun, L. Bottou, Y. Bengio, P. Haffner, Gradient-based learning applied to document recognition. *Proc. IEEE* **86**, 2278-2324 (1998).

18. J. Shi, J. Malik, Normalized cuts and image segmentation. *IEEE Trans. Pattern Anal. Mach. Intell.* **22**, 888 (2000).

19. X. Chen, Y. Duan, R. Houthooft, J. Schulman, I. Sutskever, P. Abbeel, "InfoGAN: Interpretable representation learning by information maximizing generative adversarial nets" in *Advances in*

*Neural Information Processing Systems 29*, D. D. Lee, M. Sugiyama, U. V. Luxburg, I. Guyon, R. Garnett, Eds. (Curran Associates, Inc., 2016), pp. 2172-2180.

20. A. Andoni, P. Indyk, "Near-optimal hashing algorithms for approximate nearest neighbor in high dimensions" in *47th Annual IEEE Symposium on Foundations of Computer Science, 2006. FOCS'06*, (IEEE, 2006), pp. 459-468.

21. S. Fahlman, C. Lebiere, "The cascade-correlation learning architecture" in *Advances in Neural Information Processing Systems 2*, D. S. Touretsky, Ed. (Morgan Kaufmann, San Francisco, CA, 1990), pp. 524-532.

22. S. K. Chalup, L. Wiklendt, Variations of the two-spiral task. *Conn. Sci.* **19**, 183-199 (2007).

**Acknowledgments:** The authors acknowledge the generous support of the Swiss National Science Foundation (grant no. 20021_153542/1 to R. S.) and the ETH Zurich (grant no. ETH-37 15-2 to R. S.). The authors declare no conflict of interest.

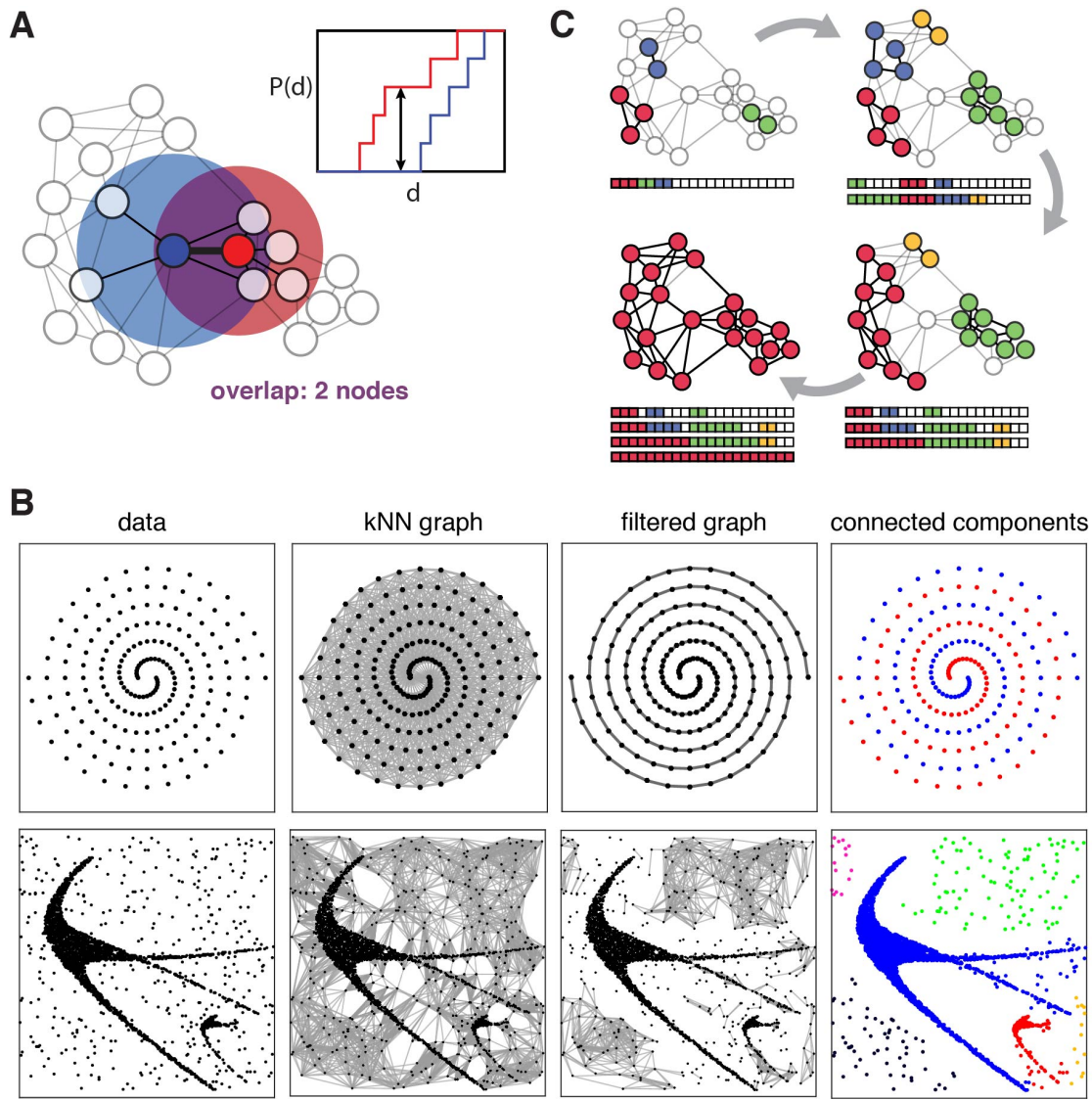

**Fig. 1. Method overview and two dimensional examples.** (**A**) Similarity calculation for edge connecting red and blue nodes, showing cumulative distributions of distance, $d$ over the neighborhoods ($k = 5$) and K-S distance (black arrow), schematic. (**B**) Two dimensional synthetic data examples *(21)* (only largest 6 connected components shown by colors, and only bidirectional edges indicated). (**C**) Adjacency matrix sorting algorithm (schematic). Rows of colored boxes indicate sort order of nodes corresponding to connected components. At each step, previous connected component orderings preserved as sub-orderings, as shown (see Supplement).

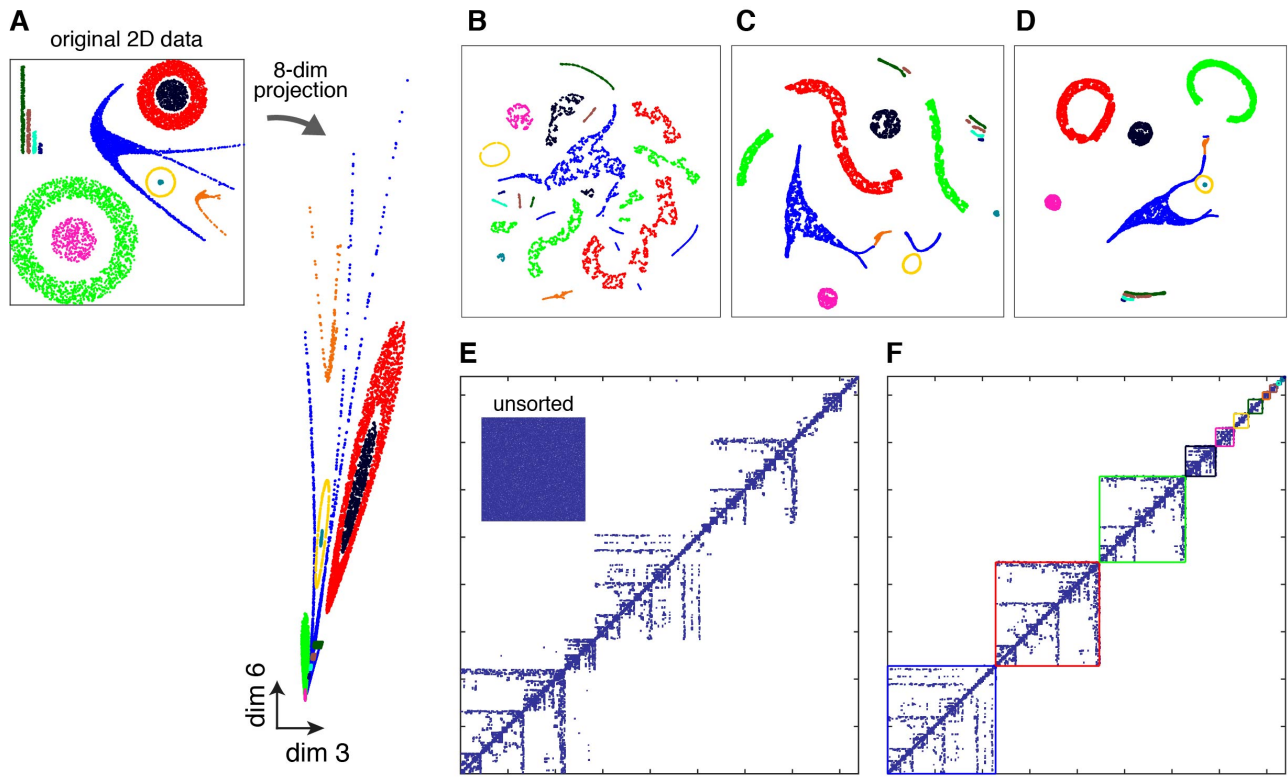

**Fig. 2. Eight dimensional synthetic example.** (**A**) Original two dimensional dataset and eight dimensional projection *(15)*. (**B-D**) t-SNE results (best of 10 runs, colors correspond to (**A**)) with t-SNE parameter perplexity set to 20, 50, 100, respectively. (**E**) *k*NN graph adjacency matrix before (inset) and after sorting. (**F**) Sorted adjacency matrix after filtering, with colored boxes indicating strongly connected components corresponding exactly to sets and colors in (**A**).

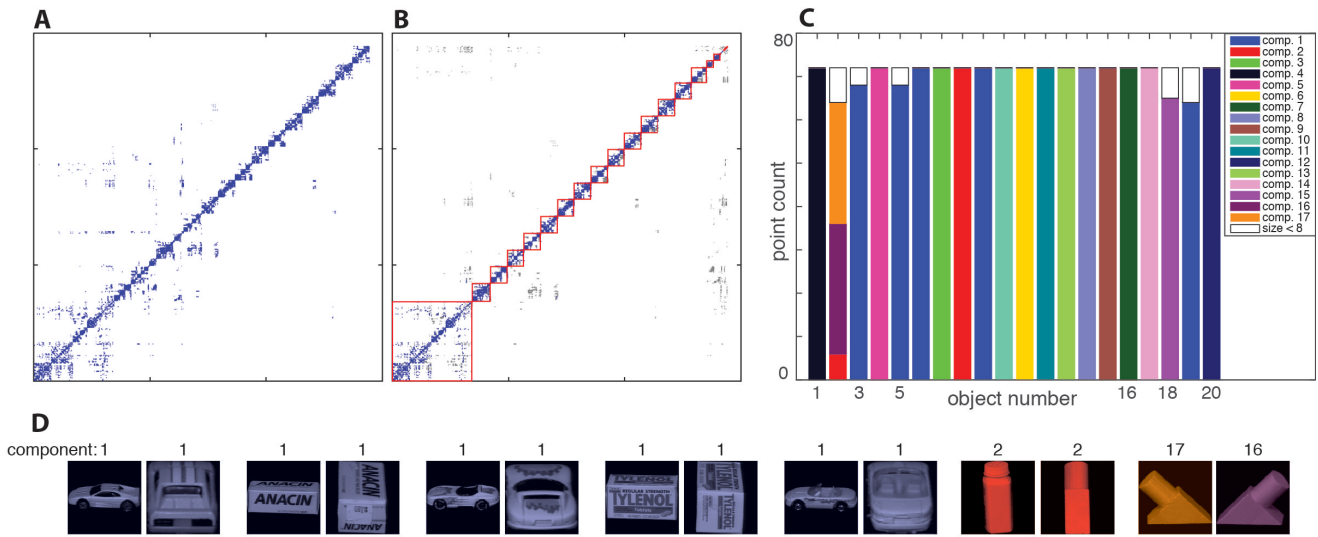

**Fig. 3. Testing on COIL-20 image dataset** *(16)*. Sorted adjacency matrix before (**A**) and after (**B**) filtering. Gray dots: removed edges; red boxes: strongly connected components. (**C**) Comparison between connected components after filtering and given object labels. (**D**) Mismatch examples.

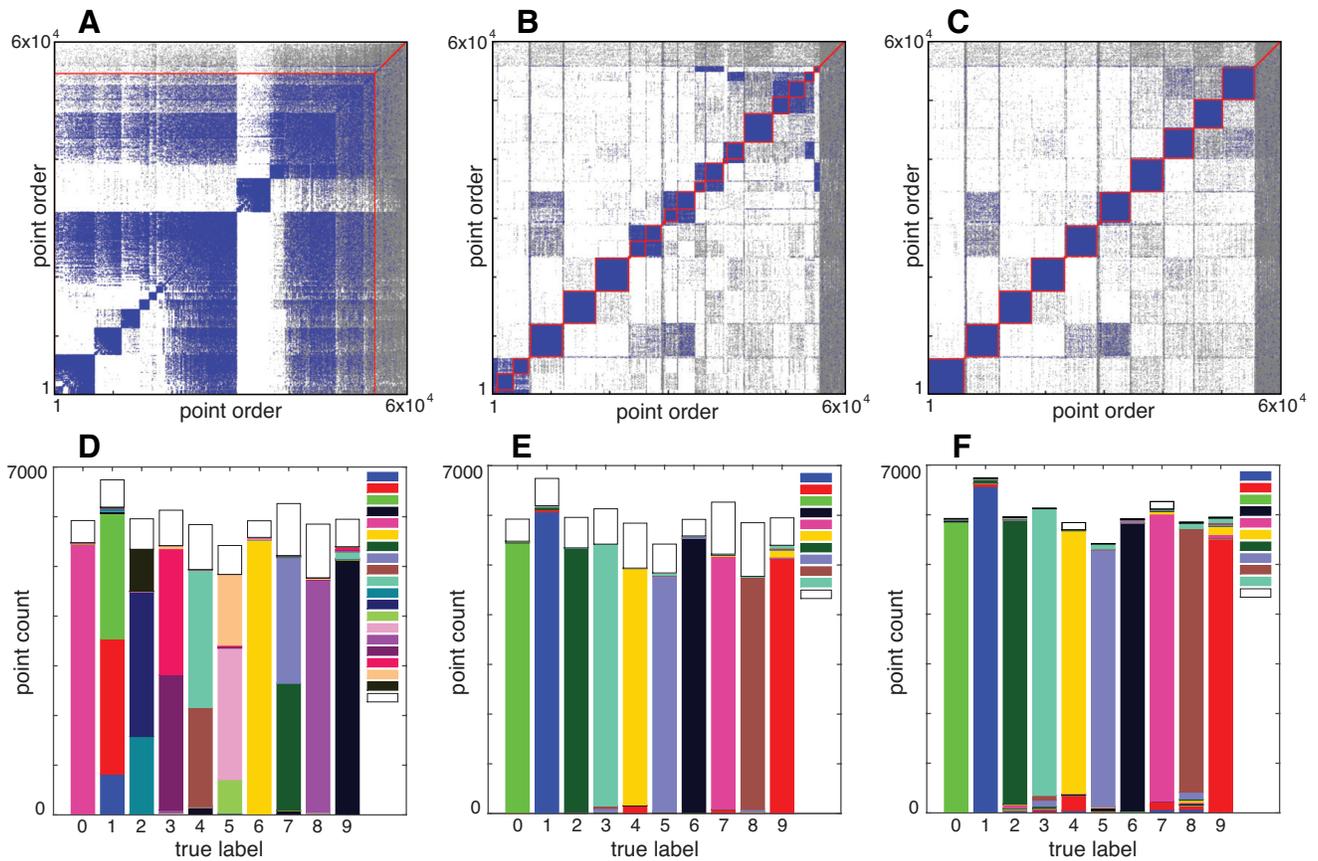

**Fig. 4. Testing on MNIST handwritten digits dataset** *(17)*. Top row: Filtered adjacency matrix (**A**), after normalized cut clustering (**B**), and cluster re-merging (**C**). Red boxes: strongly connected components or clusters; blue dots: edges preserved after filtering; gray dots: edges removed by filtering. Bottom row: Comparison between clusters (colors) and true labels, after normalized cut (**D**), after cluster re-merging (**E**), and after reassignment of points in clusters smaller than 300 points (**F**). Right-most column shows color key; clusters composed of fewer than 300 points are shown in white.

**Supplementary Materials:**

**Materials and methods:**

1. Calculation of the similarity measures

2. Adjacency matrix sorting

3. Accuracy of MNIST results

4. Spectral clustering

5. t-SNE

6. Parameter settings

7. Data details and availability

**Figs. S1 - S5**

**Movies S1 - S2 (omitted from arXiv version)**

**Materials and Methods:**

**1. Calculation of the similarity measures:** Both the K-S statistic and Jaccard similarity index take discrete values in our case, so for simplicity we refer in this supplement to discrete indices: $s_K \in \{0,...,k-1\}$, where $s_K = 0$ means the empirical distance distributions are identical, and $s_K = k - 1$ means that one node is more distant from all of its neighbors than the other; and $s_J \in \{0,...,k-1\}$, which simply counts the number of neighbors shared by the two nodes.

The combined similarity measure used in the paper for the adjacency matrix sorting is a harmonic mean of rescaled and adjusted versions of $s_K$ and $s_J$, defined as:

$$s_A = 2 \frac{\frac{s_J+1}{k}\left(1-\frac{s_K}{k+1}\right)}{\frac{s_J+1}{k}+\left(1-\frac{s_K}{k+1}\right)}.$$

**2. Adjacency matrix sorting:** A sweep of threshold values is defined between 0 and 1, with typically between 20 and 100 steps. Beginning from the largest threshold value in the sweep, at each step:

1. All edges with $s_A$ less than the threshold are removed from the $k$NN graph.

2. The strongly connected components of the resulting graph are found.

3. The points corresponding to each connected component from the *previous* step (in the first step, these are just singletons) are rearranged in blocks, such that:

    a) any previous ordering of the points within these blocks is preserved;

    b) the blocks within each new connected component are contiguous;

    c) the blocks within each new connected component are ordered by decreasing size; and

    d) the new connected components are also ordered by decreasing size.

**3. Accuracy of MNIST results:** Because the results reported on MNIST provide a partition, not a classification, we refrain from directly reporting 'accuracy'. Instead we report here the unweighted F-measure of the partitions depicted in Fig. 4. The F-measure, or $F_1$ score, compares the similarity of (partitions of) sets, and is often used in the assessment of clustering techniques. For two sets $A$ and $B$, it is defined as the binary harmonic mean between 'precision' and 'recall', where precision is the proportion of $A$ that coincides with $B$, and recall is the proportion of $B$ that coincides with $A$. To compare two sets of (sub)sets, $A_i$ and $B_j$, (e.g., the correct partition of the MNIST digits and a clustering result) we simply find the largest possible binary F-measure for each of the $A_i$ (with respect to any of the $B_j$), and average them, here using an (unweighted) mean.

F-measures:

Fig. 4 D: 0.74;

Fig. 4 E: 0.92;

Fig. 4 F: 0.95.

**4. Spectral clustering:** We follow the procedure outlined in *(18)*, to use only the second smallest eigenvector in a recursive clustering process. In addition to the stability criterion specified in that paper, we also impose a minimum cluster size, and a maximum depth of iteration (see *Parameter settings*, below).

**5. t-SNE:** We used the standard Matlab implementation of t-SNE downloaded from https://lvdmaaten.github.io/tsne/ with default settings except perplexity as indicated. For each t-SNE plot shown in the paper, t-SNE was run 10 times, and the result with the lowest value of the cost function was selected.

**6. Parameter settings:** In Fig. 1 C, the spirals were separated by removing all edges with $s_A < 0.79$, and the "shrimps" were separated by removing all edges with $s_J < 4$ or $s_K > 7$. In Fig. 2 C, all edges were removed with $s_K > 14$ or $s_J < 9$. In Fig. 3, all edges were removed with $s_K > 13$ or $s_J < 9$. In Fig. 4 A, all edges were removed with $s_K > 14$ or $s_J < 4$. In Fig 4 B, the strongly connected component of the unweighted graph from Fig. 4 A was partitioned using spectral clustering as described above, with a cut threshold of 0.1, a stability threshold of 0.04, a minimum cluster size of 50 points, and a maximum of 10 recursions permitted. In Fig. 4 C, clusters were re-merged by eye from the adjacency matrix. In Fig. 4 F, points not assigned to any of the 10 major clusters have been reassigned as follows, for each point:

1. Check whether any of the point's original $k$ nearest neighbors belongs to one of the 10 major clusters;

2. If yes, assign the point to the cluster associated with the closest of these already assigned points;

3. If no, leave the point unassigned.

This process was iterated twice to produce the result in Fig. 4 F. Some points still remain unassigned.

**7. Data details and availability:** The eight dimensional synthetic dataset is available in the supplementary material of *(15)*. The two spirals dataset in Fig. 1 B was calculated using the formulae provided in *(22)*. The 'two shrimps' dataset in Fig. 1 B is available from the authors on request. The COIL-20 dataset used was the 'processed' version downloaded from http://www.cs.columbia.edu/CAVE/software/softlib/coil-20.php . The MNIST dataset used was 'training set images' downloaded from http://yann.lecun.com/exdb/mnist/ .

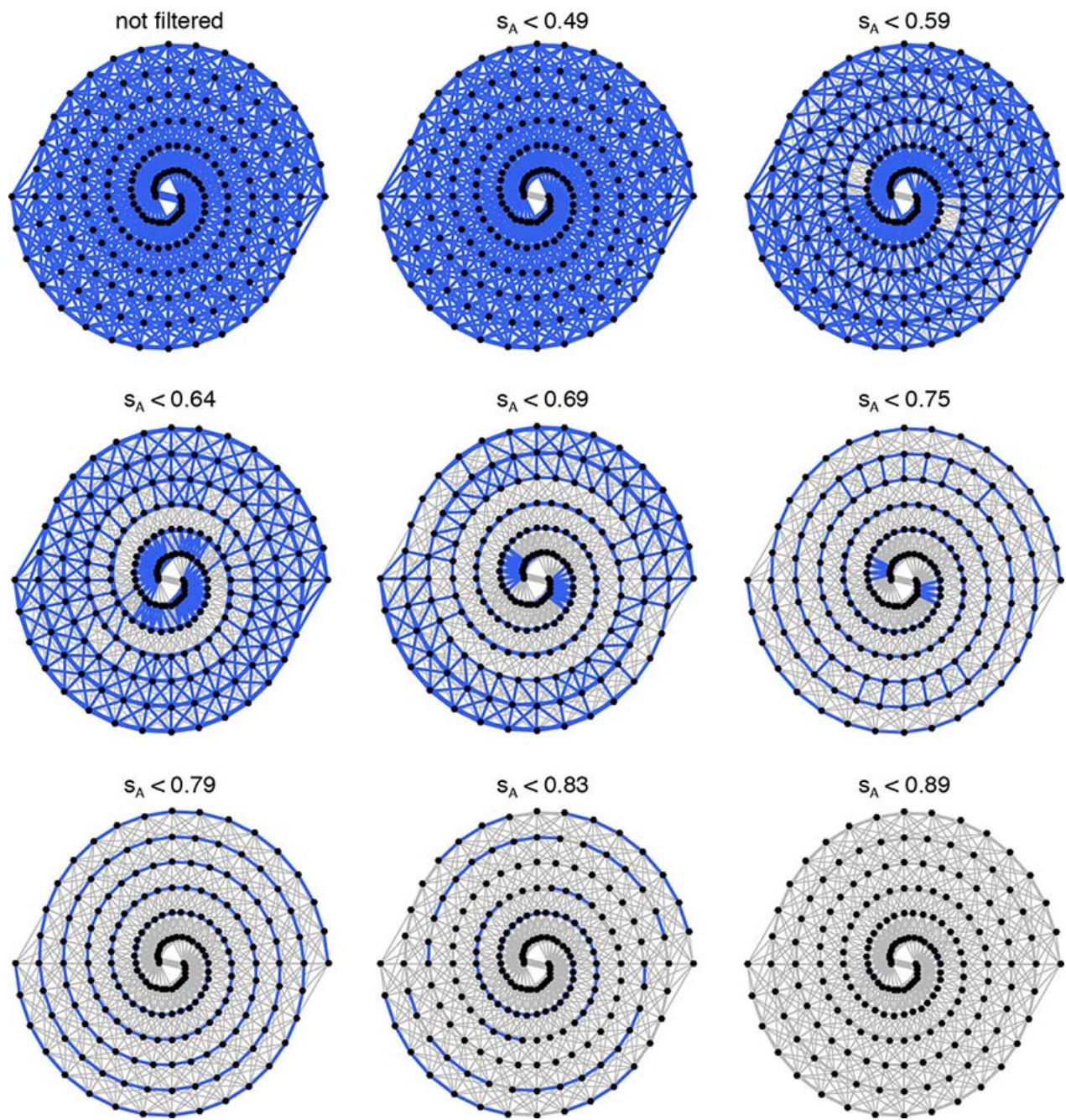

**Fig. S1:** Stability of two spirals result in Fig. 1 B. Gray: original *k*NN edges (bidirectional edges only). Blue: edges remaining after filtering (bidirectional edges only). Labels indicate edge removal criterion.

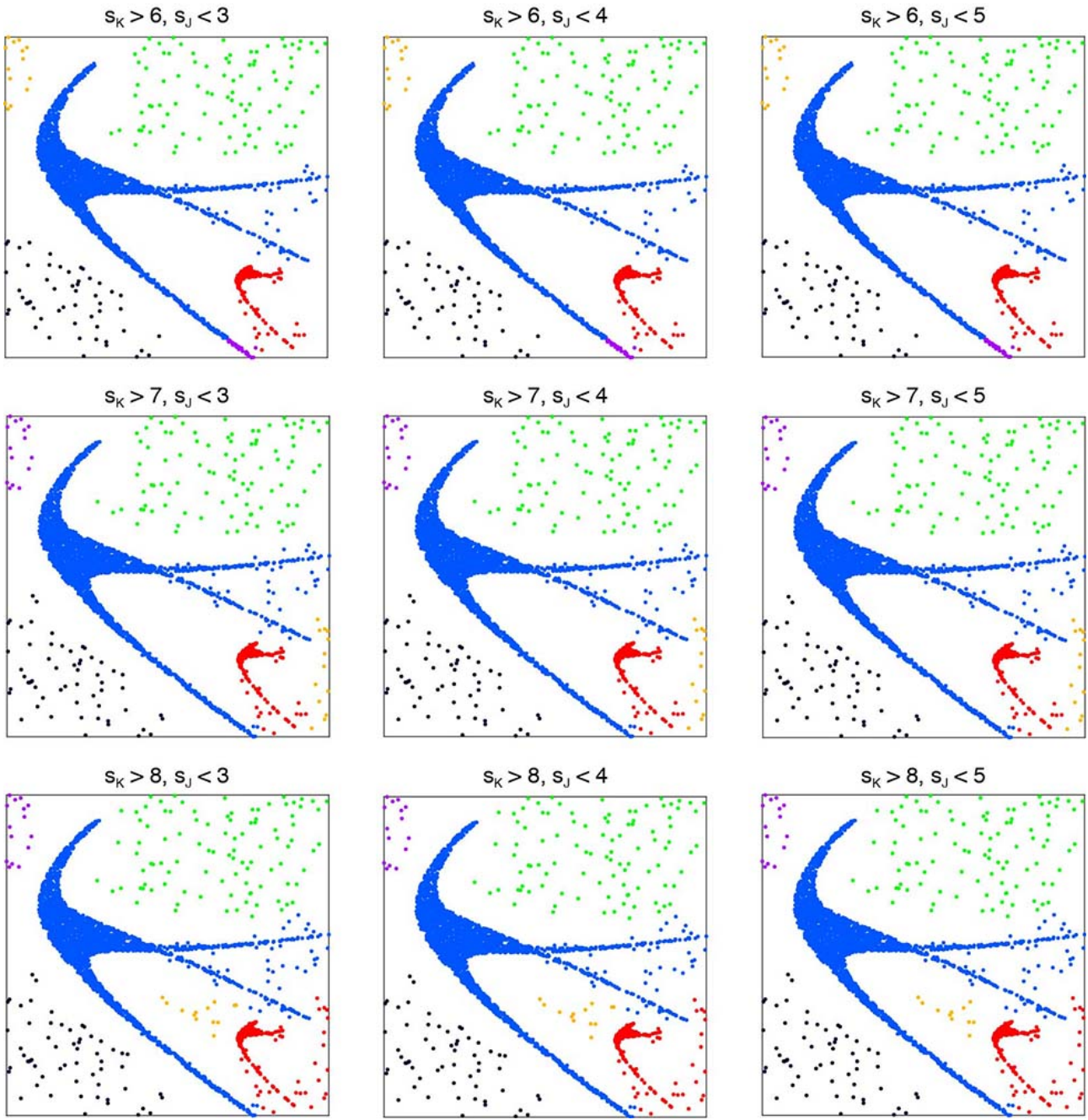

**Fig. S2:** Stability of two shrimps result in Fig. 1 B. Only six largest connected components plotted. Colors indicate six largest distinct strongly connected components of the filtered graph. Labels indicate edge removal criteria. Fig. 1 B parameter settings are in the center panel.

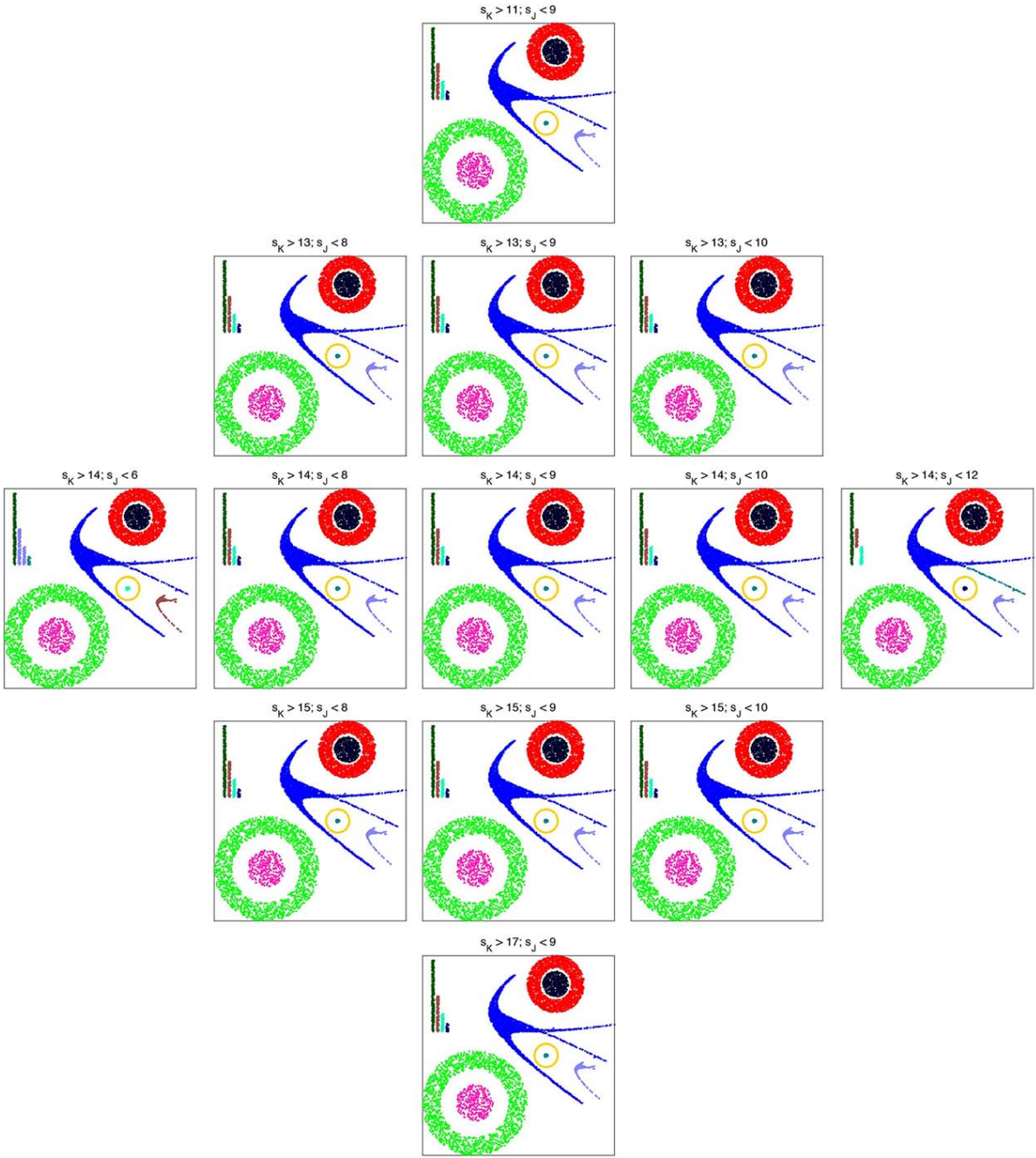

**Fig. S3:** Stability of eight dimensional dataset partition result from Fig. 2 F. Colors indicate 12 largest distinct strongly connected components. Labels indicate edge removal criteria. Procedure conducted on eight dimensional data, but results plotted in original two dimensional space for clarity. Fig. 2 F parameter settings are in the center panel.

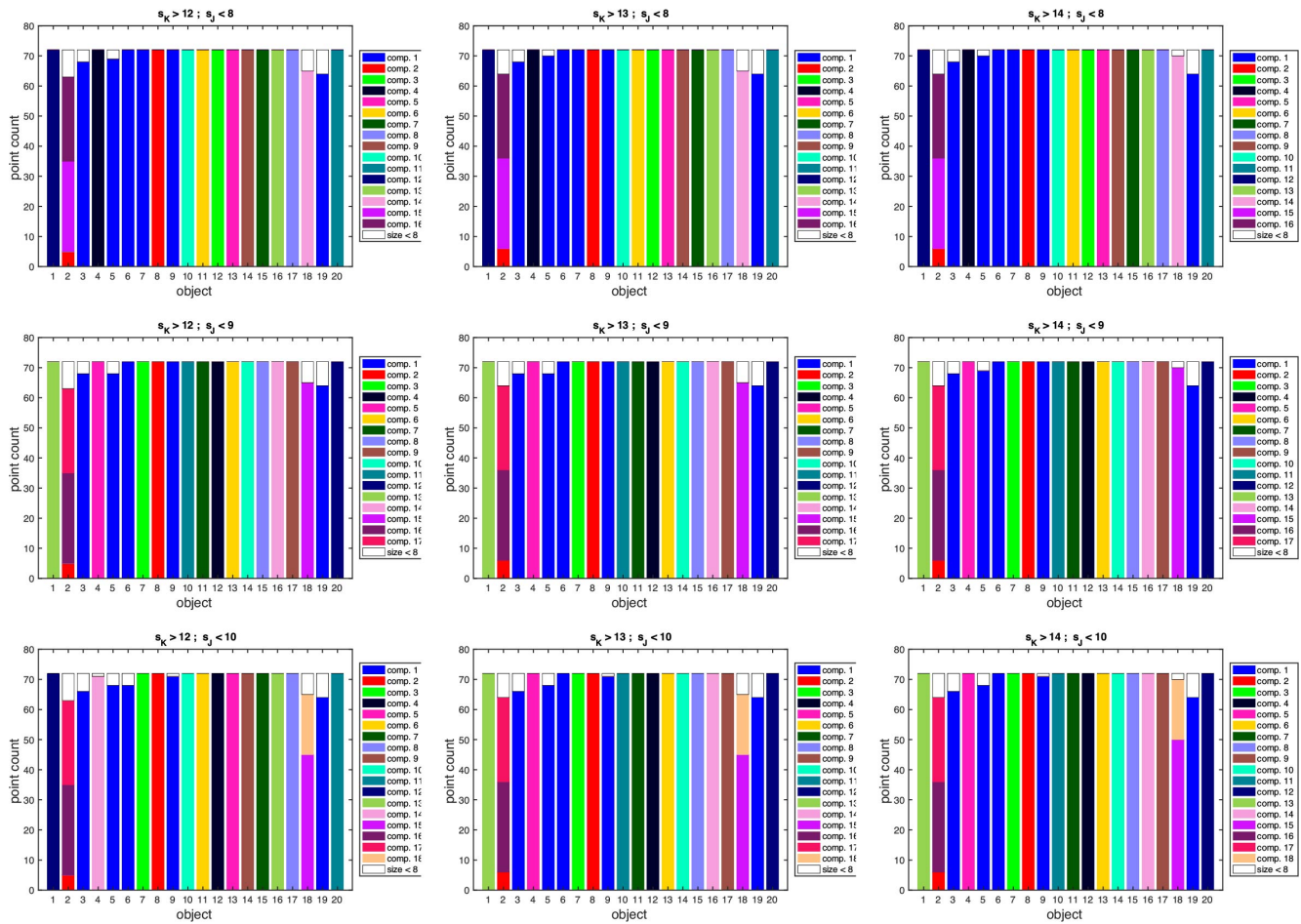

**Fig. S4:** Stability of results on COIL-20 dataset from Fig. 3 C. Labels indicate edge removal criteria. Fig. 3 C parameter settings are in the center panel.

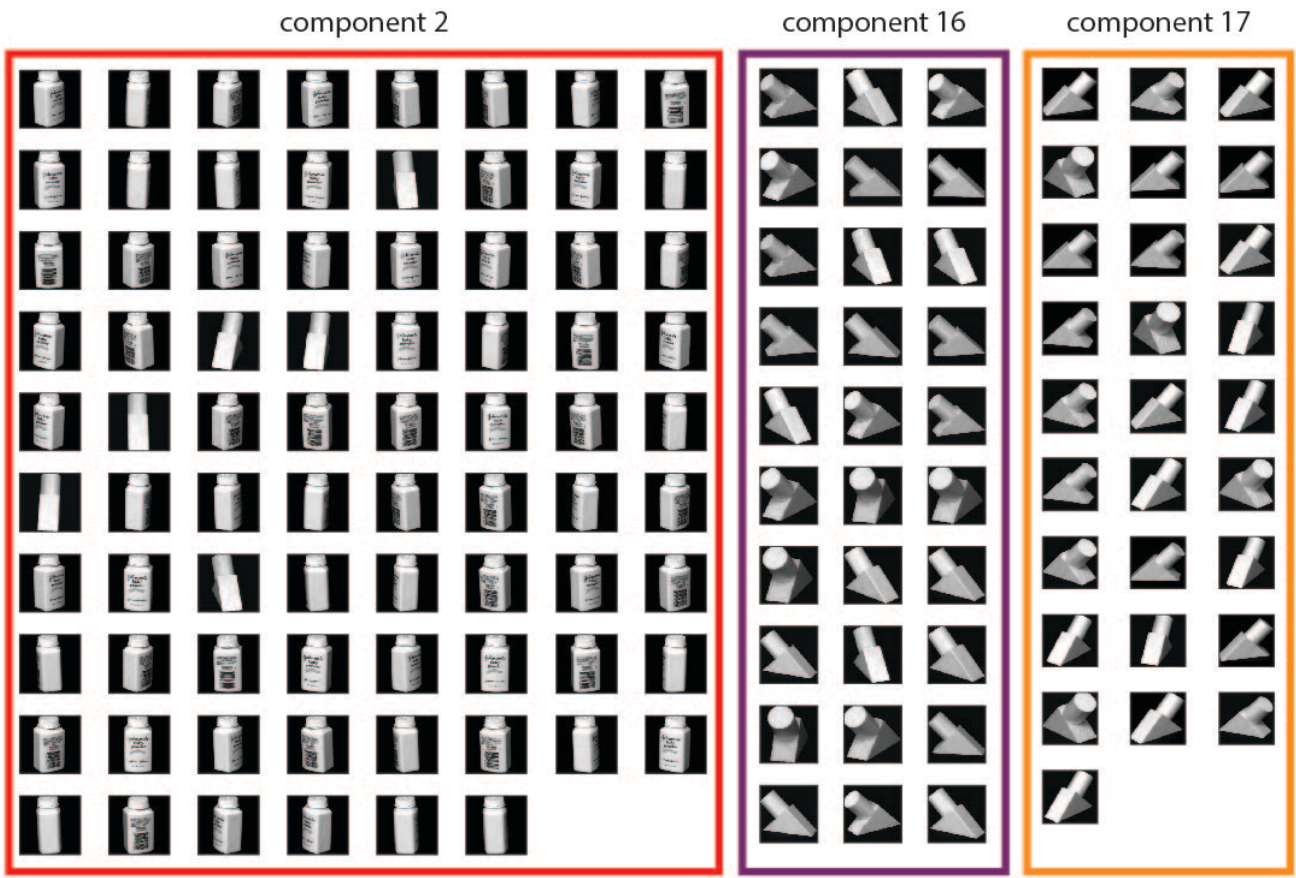

**Fig. S5:** COIL-20 images corresponding to components 2, 16 and 17 in Fig. 3.

*build8dGraph.gif*

**Movie S1:** Animation showing the sort order of nodes in the sorted adjacency matrix of Fig. 2C, and how this relates to the data.

*Shrimpgraph.gif*

**Movie S2:** Animation showing the sequence of addition of edges to the filtered graph as the filtering criterion is made more lenient on the two shrimps dataset in Fig. 1 B. Only bidirectional edges shown.